\documentclass[11pt]{article}

\usepackage[]{acl}

\usepackage{times}
\usepackage{latexsym}
\usepackage[T1]{fontenc}
\usepackage[utf8]{inputenc}
\usepackage{microtype}
\usepackage{inconsolata}  

\usepackage{amsmath,amssymb,mathtools}
\usepackage{graphicx}
\usepackage{hyperref}
\usepackage{adjustbox}
\usepackage{booktabs}
\usepackage{multirow}
\usepackage{enumitem}
\usepackage{algorithm}
\usepackage[noend]{algpseudocode}
\usepackage[skins,breakable]{tcolorbox}

\newcommand{\dd}[1]{\textsubscript{\scriptsize +#1}}
\newcommand{\ddm}[1]{\textsubscript{\scriptsize $-$#1}}

\algnewcommand{\LineComment}[1]{\Statex \(\triangleright\)~#1}

\title{Formalize Once, Edit the Rest: Efficient Lean-Based Answer Selection for Math Reasoning}

\author{
Ji Feng and Zhouxing Shi \\
University of California, Riverside \\
\texttt{\{jfeng080, zhouxing.shi\}@ucr.edu}
}

\begin{document}
\maketitle

\begin{abstract}
With large language models (LLMs) increasingly applied to mathematical reasoning, formal proof assistants such as Lean can be leveraged to verify reasoning outputs with machine-checkable rigor, enabling use cases such as answer selection in test-time scaling with $K$ sampled candidate answers. However, employing Lean requires that LLM outputs, originally in natural language, first be formalized. Existing Lean-based answer-selection work uses an autoformalization model to generate a formal statement in Lean for each candidate answer independently, incurring a significant computational cost. We propose \textsc{Base}, a base-and-edit pipeline that formalizes a single base candidate per problem and derives the remaining $K{-}1$ statements by editing the answer expression in place. To facilitate this, we train a rewriter model \textsc{LeanScribe} to localize the answer in the base formalization and generate a reusable edit function for the other $K-1$ candidates. \textsc{Base} simultaneously improves selection accuracy and reduces formalization cost---a Pareto improvement that holds on all $12$ (dataset, solver) configurations across four benchmarks and three solvers, cutting autoformalizer calls by about $5\times$ at $K{=}8$, with the reduction expected to become larger as $K$ grows. Code is available at \url{https://github.com/ucr-rai/base-and-edit}.

\end{abstract}

\section{Introduction}
\label{sec:intro}

\begin{figure*}[t]
\centering
\includegraphics[width=\textwidth]{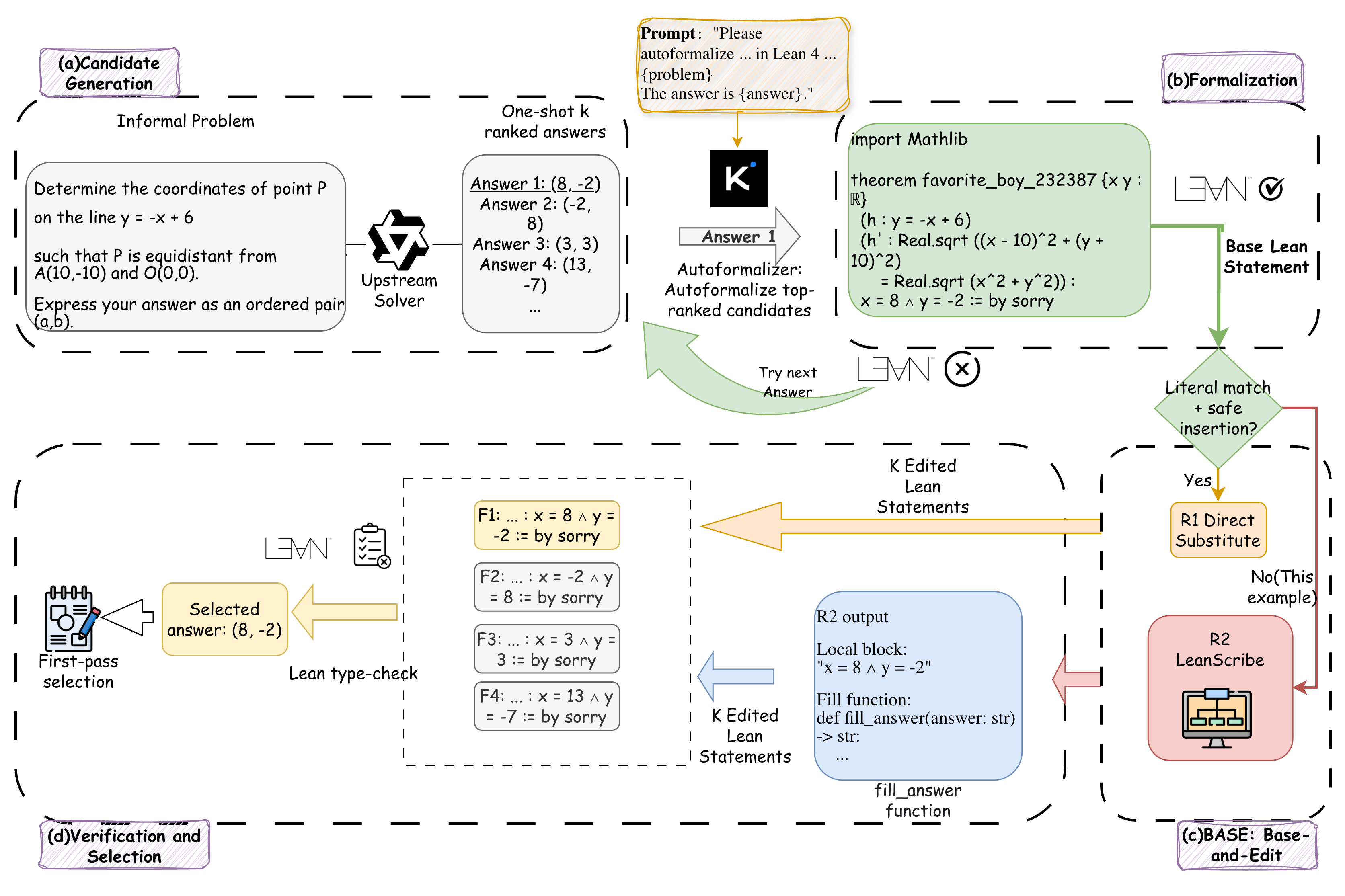}
\caption{The \textsc{Base} pipeline. An upstream solver emits $K$
ranked candidate answers; \textsc{Base} autoformalizes them in rank
order until one type-checks, giving the base statement $F_b$
(a failure triggers ``try next answer''). A two-tier
cascade---\textsc{R1} direct substitution and \textsc{R2}
\textsc{LeanScribe}---edits $F_b$ into the remaining $K{-}1$
statements, and the highest-ranked type-checking candidate is
selected.}
\label{fig:pipeline}
\end{figure*}

As large language models (LLMs) tackle increasingly hard mathematical reasoning tasks, they can often generate distinct reasoning paths for a given problem. In particular, for computational problems that require a concrete answer, as opposed to theorem-proving tasks for which only a proof is required, such reasoning paths often arrive at different answers, which raises a fundamental question: how to \emph{verify} which answer is correct? The paradigm of sampling a large number of answers and selecting among them is an instance of \emph{test-time scaling} and has driven much of the recent progress on mathematical reasoning~\citep{snell2024scaling,brown2024monkeys}.

A central challenge, however, lies in the selection step---choosing the correct answer without access to a ground-truth label. The dominant approaches are majority voting~\citep{wang2022selfconsistency} and best-of-$N$ selection with learned reward models~\citep{cobbe2021gsm8k, lightman2024prm800k}, the former relying on consensus within the candidate set and the latter on a scalar score. However, majority voting is a heuristic and fails when the correct answer is in the minority; reward models are typically black boxes. More fundamentally, both provide no formal guarantee of correctness.

To address this, a recent line of work leverages a formal verifier~\citep{zhou2024dtv,fans2024}. For each candidate answer, the problem and the answer are jointly formalized by an LLM known as an autoformalizer~\citep{kimina2025} into a  Lean~4~\citep{moura2021lean4} theorem statement asserting that the candidate is the correct answer to the problem; a prover then attempts to prove this statement~\citep{deepseekprover2024, lin2025goedelv2, liu2026numina}, with the proof verified by the Lean~4 compiler. The highest-ranked provable candidate is then selected. This paradigm is grounded in formal mathematics, as a statement with a verified proof is guaranteed to be correct.

This reliability, however, introduces a significant computational cost --- to verify $K$ candidates for a single problem, the autoformalization model is invoked $K$ times, once per candidate. 
We observe, however, that the $K$ statements for one problem are nearly identical: they typically share the same problem structure and differ only in the answer slots. A correct formalization of \emph{any one} candidate already captures this shared structure; thus the remaining $K{-}1$ statements can in principle be obtained by reusing one formalization and editing the answer expression in place.
Realizing this efficiency gain, however, is nontrivial, as candidate answers often do not appear verbatim in the Lean statement but are instead encoded in syntactically diverse forms. Therefore, reuse often requires \emph{localizing and rewriting} the answer-bearing expression rather than a simple string substitution alone.

In this work, we exploit the shared structure across a problem's candidate statements for efficient formalization, taking a step toward scalable Lean-based verification for computational problems.
Specifically, we propose \textsc{Base} (\emph{Base-And-Substitute-Edit}; \emph{base-and-edit} for short), an answer-selection pipeline that reuses a type-checked base formalization, i.e., a Lean statement accepted by the compiler as well-formed.
In this pipeline, an upstream solver first produces a list of $K$ candidate answers ranked by its own confidence---highest first. Rather than formalizing all $K$ candidate statements independently, \textsc{Base} formalizes them in rank order until one type-checks, and this type-checked base statement is then reused across the remaining candidates, deriving the remaining statements by editing only the answer expression in place.
This editing step is the technical core, handled by a two-tier cascade: a rule-based substitution for answers that appear verbatim in the base statement, and \textsc{LeanScribe}, a learned rewriter we train, for the syntactically diverse forms that rules cannot reach. \textsc{Base} then returns the highest-ranked type-checked candidate, without invoking the autoformalizer for all $K$ candidates.
In this work, \textsc{Base} focuses on efficient statement formalization and treats proof generation as orthogonal, leaving its efficiency to future work; nevertheless, statement-level type-checking already serves as an effective selection signal (Section~\ref{sec:verification}).


Figure~\ref{fig:pipeline} gives an overview and our contributions are summarized below:
\begin{enumerate}[leftmargin=*,itemsep=2pt,topsep=2pt,parsep=0pt,partopsep=0pt,label=\arabic*.]

    \item \textbf{Pipeline.}
    We propose \textsc{Base}, an answer-selection pipeline that efficiently formalizes $K$ candidates by formalizing a single base candidate and deriving the remaining $K{-}1$ statements with localized answer edits, thereby reducing the autoformalizer cost from  $\Theta(K)$ to $\Theta(1)$. To our knowledge, this is the first work to amortize formalization through a shared base.
    \textsc{Base} also serves as an efficient drop-in replacement for the per-candidate formalization stage in existing formal-selection pipelines (Section~\ref{sec:dropin}).

    \item \textbf{Model.} We introduce \textsc{LeanScribe}, a learned rewriter model trained to
    localize and re-express answers across diverse formal encodings.
    It is lightweight and invoked only once per problem, generating a reusable edit function applied to all $K{-}1$ candidates.

    \item \textbf{Experiments.} Across four benchmarks and three
    solvers, \textsc{Base} improves selection accuracy on all twelve
    (dataset, solver) configurations (up to $+23$ percentage points) while reducing
    formalization cost by roughly $5\times$---a Pareto improvement over independent formalization.

\end{enumerate}

\section{Related Work}

\paragraph{Answer selection for math reasoning.}
Step-by-step reasoning~\citep{wei2022cot} with many sampled
candidates is often adopted for math
reasoning~\citep{lewkowycz2022minerva,azerbayev2024llemma}. Without
gold labels, candidates are usually selected by majority
voting~\citep{wang2022selfconsistency} or a learned reward
model---outcome-~\citep{cobbe2021gsm8k},
process-~\citep{uesato2022process,lightman2024prm800k,wang2024mathshepherd},
or generative~\citep{zhang2024genrm}---with related work on
test-time compute allocation~\citep{snell2024scaling}.
These signals are statistical rather than formal, offering no guarantee of correctness.

\paragraph{Formal verification for mathematical reasoning.}
A growing line of work incorporates a proof assistant for mathematical reasoning.
\textsc{Fans}~\citep{fans2024} formalizes each candidate answer into a Lean~4 theorem and selects among verified ones; \textsc{Safe}~\citep{liu2025safe} applies Lean~4 verification to verify reasoning steps. Others integrate Lean into the reasoning loop~\citep{huang2025hermes} or repair proofs from compiler feedback~\citep{ospanov2025apollo}. However, these approaches formalize each statement independently, so the autoformalization cost scales linearly with the number of statements~\citep{fans2024,huang2025hermes,lu2026formalevolve}, and the shared structure across them is left unexploited.

\paragraph{Autoformalization and theorem proving.}
Autoformalization aims to translate natural-language mathematics into formal languages such as Lean~\citep{moura2021lean4}.
Existing work trains specialized models and builds pipelines for autoformalizing individual statements~\citep{lu2024forml4,kimina2025,gao2025herald,chan2025leaning,guo2025autoformalizer} and proofs~\citep{cabral2025proofflow,jana2025proofbridge}, each targeting a single statement or proof at a time.
Automated theorem provers generate or search for machine-checkable proofs of given formal statements. Modern approaches are commonly LLM-based, including next-tactic prediction~\citep{polu2020gptf,lample2022htps,yang2023leandojo} and whole-proof generation~\citep{deepseekprover2024,xin2024dspv15,lin2025goedel,xin2025bfs}.
In this work, we focus on the efficient autoformalization of statements across multiple candidates, and our pipeline builds on existing autoformalization and prover models as underlying components.


\section{Method}
\label{sec:method}

We design a pipeline for efficient Lean-based answer selection.
Given an informal mathematics problem $x$, we first elicit a ranked list of $K$ distinct candidate answers $\mathcal{S}(x) = (a_1, \ldots, a_K)$ from an upstream LLM solver $\mathcal{S}$ in a single generation call. We then formalize each problem-answer pair $(x,a_i)$  as a Lean~4 theorem statement, type-check the statements, and return the highest-ranked candidate whose verification passes (falling back to the top candidate if none does) (Section~\ref{sec:verification}).

A naive instantiation of this pipeline requires a separate autoformalizer call for each of the $K$ candidates, which we call \emph{independent formalization}.
In contrast, with our proposed \textsc{Base} (Section~\ref{sec:formalization}), we derive the remaining $K{-}1$ formal statements from a single successfully formalized and type-checked base statement, via a two-tier cascade of a rule-based rewriter and \textsc{LeanScribe}---a learned answer rewriter we train (Section~\ref{sec:leanscribe}).

\subsection{Candidate Generation}
\label{sec:generation}

The upstream solver $\mathcal{S}$ produces a \emph{ranked} list of
$K$ distinct candidate answers, ordered by the solver's own
confidence: higher-ranked candidates are those the solver believes
more likely to be correct.
We obtain this list from a single generation call for simplicity.
The selection method described below, however, is agnostic to how the candidates are produced and may be applied to candidates independently sampled through different reasoning paths.
We denote the solver's output as
\[
\mathcal{S}(x) = (a_1, a_2, \ldots, a_K),
\]
where $a_1 \prec a_2 \prec \cdots \prec a_K$ reflects the solver's
preference: $a_1$ is the most likely candidate, $a_K$ the least.
We use this ordering both during formalization and as a tiebreaker during selection.

\subsection{Formalization: Independent and Base-and-Edit}
\label{sec:formalization}

To verify a candidate answer $a_i$ with Lean, we translate the problem and that specific answer into a formal theorem statement $F_i$ using an answer-conditioned autoformalizer~$\mathcal{F}$:
\[
F_i = \mathcal{F}(x, a_i), \quad i = 1, \ldots, K.
\]
Each statement $F_i$ asserts that $a_i$ is the correct answer to $x$. We describe two strategies for generating these statements: a costly baseline (\emph{independent formalization}) and our proposed efficient approach (\emph{base-and-edit}).

\paragraph{Independent formalization (baseline).}
In this brute-force approach, $\mathcal{F}$ is invoked independently for each candidate, incurring $K$ autoformalizer calls per problem.
This matches the per-candidate strategy of prior formal answer-selection pipelines~\citep{fans2024} and serves as our efficiency reference point.

\paragraph{Base-and-edit (\textsc{Base}).}
In contrast, we propose \textsc{Base}, which replaces these $K$ independent formalizations with one formalization plus $K{-}1$ local edits. It proceeds in two steps: \emph{base discovery} and a \emph{substitution cascade}.

\emph{Base discovery.} We iterate candidates in rank order; for each $a_i$, we compute $F_i = \mathcal{F}(x, a_i)$ and check whether $F_i$ type-checks in Lean.
Let
\begin{equation}
b = \min \{\, i : F_i \text{ type-checks} \,\}.
\label{eq:bestbase}
\end{equation}
When such a $b$ exists, $F_b$ becomes the \emph{base formalization} and $a_b$ the \emph{base answer}.
Base discovery requires at most $b$ autoformalizer calls, and since recent autoformalizers~\citep{kimina2025} achieve a high compilation rate, $b$ is typically small and thus the base discovery effectively costs only $\Theta(1)$ autoformalizer calls in practice.
If no candidate type-checks, we fall back to the top-ranked candidate
$a_1$, which is rare ($1.5\%$ of problems overall; Appendix~\ref{app:abstentions}) and identical for both strategies.

\emph{Substitution cascade.} Given $(F_b, a_b)$, the cascade evaluates
each remaining candidate $a_i$ ($i \neq b$) by constructing
$F_{b \to i}$, a modified statement that reuses $F_b$'s structure with
only the answer expression replaced by $a_i$.
The cascade applies two tiers of increasing flexibility.
The first tier (\textsc{R1}) is a rule-based direct localization: it locates the base answer $a_b$ by exact string matching within the \emph{gated region} of $F_b$, where the gate is a filter that restricts matches to the theorem statement body and excludes other parts such as comments and headers (Appendix~\ref{app:gate}).
A unique match gives the substitution site directly; when $a_b$ matches at several positions, a small disambiguation model conditioned on the statement (Appendix~\ref{app:sft-details}) selects the occurrence that serves as the answer slot.
\textsc{R1} is cheap and exact, but it fails whenever the
autoformalizer does not encode the answer as a verbatim copy of
$a_b$. For these cases the second tier (\textsc{R2}) invokes
\textsc{LeanScribe}, the answer rewriter we train
(Section~\ref{sec:leanscribe}). For each candidate, the cascade
applies \textsc{R1} when direct localization and insertion are feasible,
and otherwise falls back to \textsc{R2} (Algorithm~\ref{alg:cascade}).

\begin{algorithm}[t]
\small
\caption{\textsc{Base} substitution cascade}
\label{alg:cascade}
\begin{algorithmic}[1]
\Require base statement $F_b$, base answer $a_b$, candidate $a_i$
\State $\mathcal{P} \gets \textsc{Gate}(F_b, a_b)$ \Comment{valid substitution sites for $a_b$}
\vspace{2pt}
\LineComment{\textsc{R1}: direct localization}
\If{$|\mathcal{P}| \geq 1$}
    \If{$|\mathcal{P}| = 1$}
        \State $p \gets$ unique element of $\mathcal{P}$
    \Else
        \State $p \gets \textsc{Disambig}(F_b, \mathcal{P})$
    \EndIf
    \State \Return $\textsc{Substitute}(F_b, p, a_i)$
\EndIf
\vspace{2pt}
\LineComment{\textsc{R2}: predict local block $s$ and fill function $f$}
\State $(s, f) \gets \textsc{LeanScribe}(x, F_b, a_b)$
\If{$\textsc{Gate}(F_b, s)$ is a singleton}
    \State \Return $\textsc{Substitute}(F_b, s, f(a_i))$
\EndIf
\State \Return $\bot$ \Comment{no substitution for $a_i$}
\end{algorithmic}
\end{algorithm}


\subsection{\textsc{LeanScribe}: Learned Semantic Localization and Rewriting Model}
\label{sec:leanscribe}

The forms that defeat \textsc{R1} span a broad range. For example, a
tuple ``$(3,5)$'' may be formalized as two bindings
(\texttt{a = 3}, \texttt{b = 5}) or as a single \texttt{Prod} value
(\texttt{(3, 5) : \(\mathbb{N}\) \(\times\) \(\mathbb{N}\)}), so often
no verbatim ``$(3,5)$'' substring exists to be replaced.
Similarly, intervals, sets, and symbolic forms require structural rewrites that no string match
can produce---``$[0,2)$'' becomes \texttt{Set.Ico 0 2} and
``$3\sqrt{13}$'' becomes \texttt{3 * Real.sqrt 13}.
In each case, the surrounding theorem structure binds the answer. Locating and rewriting it thus requires modeling how the autoformalizer encoded the answer in context, which is a semantic-parsing problem rather than a string-matching one.

To this end, we propose \textsc{LeanScribe}, a specialized fine-tuned LLM. Given $(x, F_b, a_b)$, it
(i)~predicts a contiguous, answer-dependent local block $s$ within $F_b$ (\emph{localization}) and
(ii)~generates a lightweight Python function $f$ mapping any informal answer to the Lean expression that replaces $s$ (\emph{transformation}).
Editing only this block preserves the verified base structure, and the reusability of $f$ allows the cost of a single model call to be amortized across all $K{-}1$ candidates: we run $f$ on each $a_i$ ($i \neq b$) to obtain $F_i \!=\! f(a_i)$, effectively replacing the answer expression of $a_b$ with that of $a_i$ while leaving the rest of $F_b$ intact.
Figure~\ref{fig:fillanswer} shows a concrete example.

\begin{figure}[t]
\scriptsize
\begin{tcolorbox}[
    enhanced,
    colback=gray!3,
    colframe=gray!50,
    boxrule=0.4pt,
    arc=2pt,
    left=4pt, right=4pt, top=2pt, bottom=2pt,
    title=\textbf{\textsc{LeanScribe} input},
    fonttitle=\footnotesize\bfseries,
    coltitle=black,
    colbacktitle=gray!15,
    attach boxed title to top left={yshift=-1mm,xshift=4mm},
    boxed title style={
        colback=gray!15,
        colframe=gray!15,
        boxrule=0pt,
        size=small,
    },
]
\textit{Problem $x$.} ``Find the distance between $(2,-6)$ and $(-4,3)$.''\\[1pt]
\textit{Base answer $a_b$.} \texttt{3{\textbackslash}sqrt\{13\}}\\[1pt]
\textit{Base formalization $F_b$.}
\begin{quote}
\ttfamily\scriptsize
theorem ex :\\
\phantom{xx}Real.sqrt ((2+4)\^{}2 + (-6-3)\^{}2)\\
\phantom{xx}\phantom{xx}= 3 * Real.sqrt 13 := by sorry
\end{quote}
\end{tcolorbox}
\vspace{-3pt}
\begin{tcolorbox}[
    enhanced,
    colback=gray!3,
    colframe=gray!50,
    boxrule=0.4pt,
    arc=2pt,
    left=4pt, right=4pt, top=2pt, bottom=2pt,
    title=\textbf{\textsc{LeanScribe} output},
    fonttitle=\footnotesize\bfseries,
    coltitle=black,
    colbacktitle=gray!15,
    attach boxed title to top left={yshift=-1mm,xshift=4mm},
    boxed title style={
        colback=gray!15,
        colframe=gray!15,
        boxrule=0pt,
        size=small,
    },
]
\textit{Local block $s$.} \texttt{"3 * Real.sqrt 13"}\\[1pt]
\textit{Fill function $g$.}
\begin{quote}
\ttfamily\scriptsize
def fill\_answer(a: str) -> str:\\
\phantom{xx}m = re.match(r'(\textbackslash d+)\textbackslash{}sqrt\{(\textbackslash d+)\}', a)\\
\phantom{xx}return f"\{m[1]\} * Real.sqrt \{m[2]\}"
\end{quote}
\textit{Inference.} For each remaining candidate $a_i$ (e.g.\ \texttt{2{\textbackslash}sqrt\{13\}}), the cascade replaces this block in $F_b$ with \texttt{fill\_answer}$(a_i)$; \textsc{LeanScribe} runs once per problem, and the synthesized function runs once per remaining candidate.
\end{tcolorbox}
\caption{\textsc{LeanScribe} prediction on a MATH-500 distance problem. The model
identifies the answer-dependent local block in $F_b$ and generates a Python function
that converts any informal answer of the same shape into the corresponding Lean expression.}
\label{fig:fillanswer}
\end{figure}

To train \textsc{LeanScribe}, we construct a training dataset via a \emph{Lean-filtered expert iteration}, inspired by prior work training reasoning models or theorem provers~\citep{zelikman2022star,lin2025goedel}: we iteratively collect model predictions, retain only those whose outputs pass the Lean type-check, and incorporate the accepted predictions into the training dataset for supervised fine-tuning (SFT).
We first build a seed dataset from \textsc{NuminaMath}~\citep{numinamath2024}: for each problem we autoformalize the gold answer with $\mathcal{F}$ to obtain a base statement, predict local blocks and fill functions using the base model, and keep only those whose reconstructed statements pass a Lean type-check.
The seed dataset is used for the first iteration of training. For subsequent iterations, using the current \textsc{LeanScribe} checkpoint, we generate block/function predictions on a broader set of problems, particularly previously failed examples, and expand the training dataset with the additional type-checked predictions. A new, stronger \textsc{LeanScribe} model is trained by SFT at each iteration as the dataset grows.

\subsection{Verification and Selection}
\label{sec:verification}

Each formal statement $F_i$, encoding a particular candidate answer $a_i$, is verified at two levels.
The \emph{statement level} involves a Lean type-check: $c_i = 1$ iff the Lean~4 compiler accepts $F_i$ as a well-formed proposition asserting $a_i$ as the answer.
The \emph{proof level} is stricter: a prover attempts a proof for the given type-checked statement, and the Lean kernel then verifies the proof, giving $p_i = 1$ iff the proof passes Lean verification.

In this work, we focus on efficient statement formalization. Proof-level verification, while more rigorous, remains too limited in coverage to serve as a practical selection signal: although current provers achieve strong performance at large pass@$K$~\citep{deepseekprover2024,lin2025goedel,liu2026numina}, their pass@1 success rates remain low (Section~\ref{sec:prover}), making proof-level selection impractical at the single-attempt budget we target. We therefore base selection on the statement-level signal $c_i$ and leave efficient proof generation to future work. A statement-level type-check certifies well-formedness of $F_i$ rather than mathematical truth of $a_i$, so selecting by $c_i$ is a practical choice that does not yet provide a correctness guarantee for the selected answer, a gap we expect stronger and more efficient provers to close in future work. Nevertheless, since the autoformalizer encodes the answer into a typed expression, a structurally inconsistent answer is more likely to yield an ill-typed statement, and combined with the solver's ranking this proves an effective selection signal in practice (Sections~\ref{sec:ablations},~\ref{sec:prover}). We study proof-level verification separately in Section~\ref{sec:prover}.

Given the signals $\{c_1, \ldots, c_K\}$ and the solver's rank order
$a_1 \prec \cdots \prec a_K$ (Section~\ref{sec:generation}), we
return the highest-ranked type-checking candidate:
\begin{equation}
\hat{a} = a_{j^*}, \qquad
j^* = \min\{\, j : c_j = 1 \,\},
\label{eq:select}
\end{equation}
falling back to $a_1$ if none type-checks.
This design combines the formal type-check signal with the solver's ranking as a tiebreaker, requiring no separate reward model.

\section{Experimental Settings}
\label{sec:experiments}

\subsection{Datasets}
\label{sec:datasets}

We evaluate on four
mathematics benchmarks requiring concrete answers,
spanning high-school olympiad to Putnam-level difficulty:
\textbf{MATH-500}~\citep{lightman2024prm800k,hendrycks2021math}, comprising the standard $500$-problem subset of MATH;
\textbf{OlympiadBench}~\citep{he2024olympiadbench}, $566$ English single-answer olympiad problems;
\textbf{AMC-AIMO}~\citep{aimo2024validation}, $83$ AMC12 problems (2022--2023) from the AIMO validation set;
and \textbf{AIME 2024}~\citep{aimo2024validation}, the $30$ problems from the 2024 AIME.
Answer types range from integers and rationals to algebraic
expressions, intervals, sets, and natural-language descriptors;
subset selection and answer normalization are detailed in
Appendix~\ref{app:datasets}.

\subsection{Models and Implementation}
\label{sec:models-impl}

To assess generality across solver families, we use three upstream solvers for generating candidates:
\textbf{Gemini~3 Pro},
\textbf{Qwen3-8B}~\citep{yang2025qwen3}, and
\textbf{DeepSeek-R1-0528-Qwen3-8B}~\citep{deepseekr1}
(\emph{R1-Qwen3-8B} in tables), all at $K{=}8$.
For verification, the autoformalizer~$\mathcal{F}$ is
\textbf{Kimina-Autoformalizer-7B}~\citep{kimina2025};
type-checking uses the Lean~4 compiler~\citep{moura2021lean4},
and the proof step uses \textbf{DeepSeek-Prover-V2-7B}~\citep{deepseekprover2024} with a single greedy attempt per type-checked statement.
For the learned rewriter \textsc{LeanScribe} (Section~\ref{sec:leanscribe}), it is based on Qwen3-8B~\citep{yang2025qwen3} fine-tuned via LoRA on the expert-iterated training dataset with details in Appendix~\ref{app:sft-details}.
All inference runs on a single RTX~PRO~6000 GPU with vLLM and greedy decoding, and
Lean~4~\citep{moura2021lean4} verification uses Mathlib~4~\citep{mathlib2020} with a $60$s per-statement timeout.

\subsection{Baseline and Metrics}
\label{sec:baseline}

\paragraph{Baseline.}
We compare primarily against independent formalization (Section~\ref{sec:formalization}):
the same pipeline but with $K$ independent autoformalizer calls,
instead of one base call plus $K{-}1$ local edits.
We use identical verifier components, prompts, and selection rule for both methods, so
any difference is attributable to the formalization strategy.

\paragraph{Metrics.} All main results use the statement-level
signal (type-check); the proof level is studied separately
(Section~\ref{sec:prover}), and ``verification'' means the statement
level unless noted.
\textbf{Acc} is the fraction of problems whose selected answer equals the gold answer.
\textbf{GT@K} is the fraction whose gold answer appears in the ranked list
(determined solely by the upstream solver and thus identical for both methods),
and \textbf{Acc\,|\,GT@K} is \textsc{Acc} on that subset, isolating verifier quality;
thereby, $\mathrm{Acc} = \mathrm{GT}@K \cdot \mathrm{Acc}\,|\,\mathrm{GT}@K$.
For verifier diagnostics,
\textbf{GT-pass} is the fraction whose gold statement type-checks and
\textbf{P-cand} the average fraction of candidate statements that type-check (ablation only).
For computational costs, we report \textbf{autoformalizer calls} per problem.

\section{Main Results}
\label{sec:main-results}

\subsection{Accuracy}
\label{sec:accuracy}

Table~\ref{tab:main} reports our main result: \textsc{Base}'s
end-to-end accuracy (\textsc{Acc}) and gold-pass rate (GT-pass)
across four datasets and three upstream solvers, against the
independent formalization baseline. The pattern is consistent across
solvers and datasets. \textsc{Base} exceeds independent
formalization on \textsc{Acc} in all twelve (dataset, solver) cells,
with gains from $+0.7$ to $+23.3$ points. It
also raises GT-pass in all twelve cells (up to $+33.3$ points,
remaining high across all cells, $96$--$100\%$): the gold candidate's statement
almost always survives under \textsc{Base}, so sharing a verified
base makes the gold formalization markedly more robust.

\begin{table*}[t]
\centering
\small
\setlength{\tabcolsep}{8pt}
\begin{tabular}{l l | c c | c c}
\toprule
& & \multicolumn{2}{c|}{\textsc{Acc} (\%)} &
    \multicolumn{2}{c}{GT-pass (\%)} \\
Dataset & Solver & Indep. & \textsc{Base} (ours) & Indep. & \textsc{Base} (ours) \\
\midrule
\multirow{3}{*}{MATH-500}
    & Gemini 3 Pro & 70.2 & \textbf{79.4}\dd{9.2}  & 87.3 & \textbf{98.7}\dd{11.4} \\
    & Qwen3-8B     & 45.7 & \textbf{51.2}\dd{5.5}  & 88.2 & \textbf{96.3}\dd{8.1}  \\
    & R1-Qwen3-8B  & 58.7 & \textbf{67.5}\dd{8.8}  & 85.4 & \textbf{98.8}\dd{13.4} \\
\midrule
\multirow{3}{*}{OlympiadBench}
    & Gemini 3 Pro & 41.7 & \textbf{53.4}\dd{11.7} & 77.6 & \textbf{99.0}\dd{21.4} \\
    & Qwen3-8B & 31.6 & \textbf{39.7}\dd{8.1}  & 80.4 & \textbf{98.1}\dd{17.7} \\
    & R1-Qwen3-8B & 42.8 & \textbf{53.2}\dd{10.4} & 79.4 & \textbf{100.0}\dd{20.6} \\
\midrule
\multirow{3}{*}{AMC-AIMO}
    & Gemini 3 Pro & 73.5 & \textbf{86.7}\dd{13.2} & 82.9 & \textbf{100.0}\dd{17.1} \\
    & Qwen3-8B & 37.8 & \textbf{46.8}\dd{9.0}  & 72.9 & \textbf{97.9}\dd{25.0} \\
    & R1-Qwen3-8B & 67.5 & \textbf{85.0}\dd{17.5} & 81.2 & \textbf{100.0}\dd{18.8} \\
\midrule
\multirow{3}{*}{AIME 2024}
    & Gemini 3 Pro & 40.0 & \textbf{63.3}\dd{23.3} & 66.7 & \textbf{100.0}\dd{33.3} \\
    & Qwen3-8B & 20.0 & \textbf{20.7}\dd{0.7}  & 85.7 & \textbf{100.0}\dd{14.3} \\
    & R1-Qwen3-8B & 44.8 & \textbf{60.7}\dd{15.9} & 73.7 & \textbf{100.0}\dd{26.3} \\
\bottomrule
\end{tabular}
\caption{Main results across four datasets and three upstream
solvers ($K{=}8$). \emph{Indep.}${=}$independent formalization.}
\label{tab:main}
\end{table*}

\subsection{Cost}
\label{sec:cost}

Table~\ref{tab:cost} gives the per-problem formalization cost. The
autoformalizer dominates pipeline cost and is where the two
pipelines differ in \emph{scaling}: the baseline issues one call per
candidate ($\Theta(K)$), while \textsc{Base} issues one base call
plus at most one substitution call, giving $\Theta(1)$ cost determined
by the base rank $b\,({\approx}1.5)$. The substitution call invokes
\textsc{LeanScribe} once per problem to predict a reusable edit
function (Section~\ref{sec:leanscribe}); this single prediction is
applied to all $K{-}1$ candidates and is far cheaper than the
$K{-}1$ autoformalizer calls it replaces, so it does not affect the
$\Theta(1)$ scaling. At $K{=}8$ this is
$b/K{=}0.19\times$ of independent formalization---a $4.7$--$6.5\times$
reduction (mean $5.4\times$)---and the gap widens with $K$.
Together with the accuracy result of Section~\ref{sec:accuracy}, \textsc{Base} is strictly cheaper while never less accurate across all twelve configurations, achieving a Pareto improvement.

\begin{table}[t]
\centering
\small
\setlength{\tabcolsep}{4pt}
\begin{adjustbox}{max width=\columnwidth}
\begin{tabular}{l c c c}
\toprule
Dataset & Gemini 3 Pro & Qwen3-8B & R1-Qwen3-8B \\
\midrule
MATH-500      & 1.23 & 1.29 & 1.23 \\
AMC-AIMO      & 1.49 & 1.65 & 1.43 \\
AIME 2024     & 1.67 & 1.50 & 1.70 \\
OlympiadBench & 1.65 & 1.57 & 1.45 \\
\bottomrule
\end{tabular}
\end{adjustbox}
\caption{Per-problem formalization cost of \textsc{Base}  (average number of base-discovery calls). In contrast, the independent formalization baseline requires $K{=}8$ autoformalizer calls per problem.}
\label{tab:cost}
\end{table}

\subsection{Ablations}
\label{sec:ablations}


\begin{table}[t]
\centering
\adjustbox{max width=.42\textwidth}{
\begin{tabular}{lcccc}
\toprule
Dataset & Indep. & \textsc{Base} & $-$R2 & $\Delta_{\text{R2}}$ \\
\midrule
MATH-500      & 87.5 & 87.9 & 67.9 & \textbf{+20.0} \\
AMC-AIMO      & 76.6 & 93.9 & 85.6 & \textbf{+8.3}  \\
AIME 2024     & 76.1 & 82.6 & 60.1 & \textbf{+22.5} \\
OlympiadBench & 76.8 & 86.4 & 67.7 & \textbf{+18.7} \\
\bottomrule
\end{tabular}}
\caption{Component ablation (P-cand \%, Gemini, $K=8$).
``Indep.'' is the baseline;
\textsc{Base} uses both R1 and R2;
``$-$R2'' disables R2, leaving R1 only;
``$\Delta_{\text{R2}}$'' indicates the
contribution of R2 on top of using R1 only.}
\label{tab:ablation}
\end{table}

Table~\ref{tab:ablation} isolates the contribution of
\textsc{LeanScribe}, the learned answer rewriter (\textsc{R2}).
The ablation study is performed on MATH-500 (Gemini~$K{=}8$), measured by the per-candidate Lean pass rate (P-cand)---the fraction of all candidate statements that verify.
Removing \textsc{R2} leaves only the rule-based localization tier (\textsc{R1}).
\textsc{LeanScribe} accounts for $8$ to $22$ points of per-candidate
coverage ($\Delta_{\textsc{R2}}$). Rule-only localization
(\textsc{R1}) handles answers appearing as verbatim substrings of
the base, but \textsc{LeanScribe} recovers many candidates that
encode answers differently; the two tiers are thus complementary,
with \textsc{R1} covering easy verbatim edits and \textsc{R2}
the harder cases.

\textsc{LeanScribe} also generalizes out of
distribution: although fine-tuned only on NuminaMath
(Section~\ref{sec:models-impl}), it attains high GT-pass on every
benchmark (Table~\ref{tab:main}, $96$--$100\%$), indicating that its
localization-and-rewriting behavior transfers rather than memorizing
NuminaMath patterns, with residual failures concentrated on very long
answer expressions and unusual geometric constructions.

\subsection{Proof-level analysis}
\label{sec:prover}

We also run the full pipeline that further consists of proving the formal statements with a prover, and we first use \textsc{DeepSeek-Prover-V2}~\citep{deepseekprover2024}.
This studies whether the statement-level gains carry through to a prover that must close each theorem.
Prior work on provers typically reports pass@$k$ with a large $k$, sampling many proof attempts per theorem.
In our setting, however, each of the $K$ candidate answers needs its own proof, so a large $k$ for sampling proofs would multiply the prover cost by a factor of $K$ and is impractical.
We therefore use a single pass@1 proof attempt per candidate.

\begin{table}[t]
\centering
\small
\setlength{\tabcolsep}{5pt}
\begin{adjustbox}{max width=\columnwidth}
\begin{tabular}{l | c c c}
\toprule
Dataset & Gemini 3 Pro & Qwen3-8B & R1-Qwen3-8B \\
\midrule
MATH-500      & 31.4\dd{3.9} & 25.9\dd{3.2} & 28.4\dd{3.2} \\
AMC-AIMO      & 13.3\dd{3.5} & 8.9\dd{1.3}  & 16.2\dd{5.0} \\
AIME 2024     & 6.7\dd{3.4}  & 6.9\dd{3.5}  & 3.6\ddm{7.1} \\
OlympiadBench & 11.9\dd{2.8} & 11.2\dd{2.5} & 11.8\dd{3.3} \\
\bottomrule
\end{tabular}
\end{adjustbox}
\caption{Proof-level accuracy (\%) for \textsc{Base}, compared against the independent formalization baseline with gains denoted in the subscripts. We use $K{=}8$ for candidates and DeepSeek-Prover-V2 for the prover.}
\label{tab:prover}
\end{table}

Table~\ref{tab:prover} shows the results.
We notice that the proof-level accuracies are low ($4$--$31\%$) compared to results in Table~\ref{tab:main}, which is because of the limited capability of the prover under the single-attempt budget.
Nevertheless, we still compare \textsc{Base} against independent formalization in each setting, which is the focus of this work.
\textsc{Base} improves proof-level accuracy on 11 out of
12 cells (sign test $p{=}0.006$), the lone exception being the
smallest cell (AIME-2024, R1-Qwen3-8B, $n{=}28$). This echoes prior
findings that the quality of the formal statement shapes how a
downstream prover fares~\citep{jiang2023dsp,wu2022autoformalization}.
The per-candidate false-positive rate---an incorrect answer
receiving a complete proof---is a low ${\sim}7\%$ on average
($1$--$14\%$ across cells), arising when the autoformalizer renders an incorrect answer as a statement that type-checks and admits a proof.
Given such false positives, we combine the formal signal with the solver's ranking rather than relying on a single proof.

\paragraph{Swapping the prover.}
\textsc{Base} feeds type-checked candidates to an off-the-shelf
prover, so the prover should be interchangeable. Replacing
\textsc{DeepSeek-Prover-V2} with
\textsc{Kimina-Prover-Preview-Distill-7B}~\citep{kimina2025}
(Table~\ref{tab:prover-swap}), the two are close on MATH-500 but
\textsc{DeepSeek-Prover-V2} is about twice as strong on AMC-AIMO. The
advantage is dataset-dependent, and base-and-edit is agnostic to the
prover used.

\begin{table}[t]
\centering
\small
\setlength{\tabcolsep}{6pt}
\begin{adjustbox}{max width=\columnwidth}
\begin{tabular}{l l c c}
\toprule
Dataset & Prover & Complete (\%) & GT-complete (\%) \\
\midrule
\multirow{2}{*}{MATH-500}
   & DSP-V2 & \textbf{7.3}\dd{0.8} & \textbf{31.4}\dd{2.3} \\
   & Kimina & 6.5 & 29.1 \\
\midrule
\multirow{2}{*}{AMC-AIMO}
   & DSP-V2 & \textbf{8.1}\dd{3.9} & \textbf{14.5}\dd{6.1} \\
   & Kimina & 4.2 & 8.4 \\
\bottomrule
\end{tabular}
\end{adjustbox}
\caption{Prover swap on \textsc{Base}-formalized candidates.
\emph{Complete}/\emph{GT-complete}: candidate-level and gold-answer
proof-completion rates.}
\label{tab:prover-swap}
\end{table}

\subsection{Drop-in for an Existing Pipeline}
\label{sec:dropin}

\begin{table}[t]
\centering
\small
\adjustbox{max width=\textwidth}{
\begin{tabular}{l c c}
\toprule
Dataset & Calls (Indep.$\to$\textsc{Base}) & Acc \\
\midrule
MATH-500  & $4000 \to 689$~{\scriptsize($5.8\times$ $\downarrow$)} & $88.2 \to 88.6$ \\
AMC-AIMO  & $664 \to 116$~{\scriptsize($5.7\times$ $\downarrow$)} & $62.7 \to 61.4$ \\
AIME 2024 & $240 \to 53$~{\scriptsize($4.5\times$ $\downarrow$)} & $13.3 \to 13.3$ \\
\bottomrule
\end{tabular}}
\caption{Dropping \textsc{Base} into a \textsc{Fans}-style
majority-voting pipeline (Qwen2.5-Math-7B-Instruct, $K{=}8$) by replacing only the formalization stage. \emph{Calls}: autoformalizer
calls per dataset; \emph{Acc}: selection accuracy (\%).}
\label{tab:dropin}
\end{table}

Our \textsc{Base} pipeline described in Section~\ref{sec:method} is streamlined to focus on the formalization stage and enable efficient formalization for $K$ candidates. In practice, Lean-based answer selection can be more sophisticated and can contain proving statements in addition to formalization. Our efficient base-and-edit formalization in \textsc{Base} can be seamlessly integrated into other Lean-based answer-selection pipelines.
We demonstrate this with \textsc{Fans}~\citep{fans2024}, which generates proofs for statements, checks consistency between natural language and formal language with a QwQ-32B judge, and enhances the selection by majority vote.
We use candidate answers from  Qwen2.5-Math-7B-Instruct, following \textsc{Fans}, and sample $K{=}8$ candidates. Keeping other components fixed, we replace only the formalization stage:
\textsc{Fans} originally uses independent formalization;
in contrast, we derive one Lean-verified base statement and edit it for the rest.
In this way, we isolate the impact of our improved autoformalization, and compare the number of autoformalizer calls and final selection accuracy under an otherwise identical downstream pipeline.
Table~\ref{tab:dropin} shows the results, with experimental details in Appendix~\ref{app:dropin}.
Our \textsc{Base} cuts autoformalizer calls by $4.5$--$5.8\times$ while maintaining accuracy. This demonstrates that \textsc{Base} is a flexible and efficient module that can be integrated into other pipelines, such as \textsc{Fans}, that make contributions orthogonal to efficient autoformalization for Lean-based answer-selection.

\section{Mechanism and Analysis}
\label{sec:analysis-mech}

\paragraph{Factorizing the gain.}
\label{sec:analysis}

End-to-end accuracy factorizes cleanly into
\begin{equation}
\mathrm{Acc} = \mathrm{GT}@K \cdot \mathrm{Acc}\,|\,\mathrm{GT}@K,
\label{eq:factorization}
\end{equation}
where the first factor depends mostly on the upstream solver
(both methods score the same candidate list) and the second is the
verifier's conversion rate on solver-correct problems, so any sizable
difference in $\mathrm{Acc}$ comes from the verifier.
Table~\ref{tab:factorization} confirms this on MATH-500:
$\mathrm{GT}@K$ is essentially identical across methods, while
\textsc{Base}'s $\mathrm{Acc}\,|\,\mathrm{GT}@K$ is uniformly higher
across all three solvers. The gain is thus a
verifier effect: \textsc{Base} does not change which answers the
solver proposes, but raises the rate at which a solver-correct answer
is \emph{retained}.

\begin{table}[!t]
\centering
\small
\setlength{\tabcolsep}{4pt}
\begin{tabular}{l l c c c}
\toprule
Solver & Method & $\mathrm{GT}@K$ & $\mathrm{Acc}\,|\,\mathrm{GT}@K$ & $\mathrm{Acc}$ \\
\midrule
\multirow{2}{*}{Gemini 3 Pro} & Indep. & 80.4 & 87.3 & 70.2 \\
                              & \textsc{Base} & 80.6 & \textbf{98.5} & \textbf{79.4} \\
\midrule
\multirow{2}{*}{Qwen3-8B}     & Indep. & 60.9 & 75.0 & 45.7 \\
                              & \textsc{Base} & 60.7 & \textbf{84.3} & \textbf{51.2} \\
\midrule
\multirow{2}{*}{R1-Qwen3-8B}  & Indep. & 69.0 & 85.1 & 58.7 \\
                              & \textsc{Base} & 68.8 & \textbf{98.2} & \textbf{67.5} \\
\bottomrule
\end{tabular}
\caption{Factorization on MATH-500 (\%).}
\label{tab:factorization}
\end{table}

\paragraph{Where the gain comes from: the shared verified base.}
\label{sec:mech-ablation}

\begin{table}[t]
\centering
\small
\setlength{\tabcolsep}{4pt}
\begin{tabular}{l c c c c}
\toprule
Setting & GT-pass & P-cand & Acc & Acc\,$|$\,GT@K \\
\midrule
Indep.\ baseline       & 81.9 & 81.2 & 39.3 & 66.7 \\
First verified base  & \textbf{98.2} & \textbf{91.7} & \textbf{47.1} & \textbf{79.5} \\
Random verified base & 95.4 & 91.2 & 45.6 & 77.0 \\
\bottomrule
\end{tabular}
\caption{Ablating the shared verified base (\%) on a matched set of
OlympiadBench, AMC-AIMO, and AIME~2024.}
\label{tab:base-ablation}
\end{table}

To trace the gain to the shared verified base, we compare three
strategies on the same candidate lists (Table~\ref{tab:base-ablation}).
Both variants that reuse a single verified base---\textsc{Base},
which reuses the \emph{first} type-checking candidate, and a control
that reuses a \emph{random} one---outperform independent
formalization, and crucially \emph{first~$\approx$~random}: the
benefit comes from structural reuse, not from cherry-picking a good
formalization.
This traces to one mechanism---sharing a single verified scaffold
rather than formalizing independently. Statistically, sharing cuts
per-candidate formalization noise ($6.3\times$ lower pairwise
variance, though only part of the story: $r{=}0.22$ with per-problem
gains); mechanistically, every candidate inherits the base's verified
structure rather than risking its own autoformalizer failure. Either
way, sharing raises the type-check rate by $+9.9$ points
(Appendix~\ref{app:funnel}).

\section{Conclusion}

We present \textsc{Base}, a base-and-edit approach to formal
answer selection that formalizes a single verified base per problem
and derives the remaining candidates by editing only the answer
block. Replacing $K$ independent autoformalizer calls with one base
call plus localized edits cuts formalization cost from $\Theta(K)$
to $\Theta(1)$ while improving selection accuracy---a
Pareto improvement on all twelve configurations.
Our analysis traces the gain to the verifier and the use of shared base formalization.
We believe this work will enable more practical answer-selection and verification for natural-language reasoning with Lean.

\section*{Acknowledgement}
This work is supported in part by an NVIDIA Academic Grant Program award.

This work used the Delta system at the National Center for Supercomputing Applications [award OAC 2005572] through allocation CIS250704 from the Advanced Cyberinfrastructure Coordination Ecosystem: Services \& Support (ACCESS) program, which is supported by National Science Foundation grants \#2138259, \#2138286, \#2138307, \#2137603, and \#2138296.

\section*{Limitations}

This work focuses on the efficiency of statement autoformalization for $K$ candidate answers, and we use statement-level type-checking as the selection signal due to the limitation of provers. Future work is needed to address the efficiency of provers when handling $K$ relevant theorem statements for the candidate answers, and enable an efficient answer-selection with both statement autoformalization and theorem proving. 
In our method, base-and-edit assumes the $K$ candidate answers share problem-side structure so that a single base formalization can be reused; this holds for the tasks we study but may not transfer to settings where each candidate requires a structurally different statement.
Lastly, since we find that prover can appear to complete proofs for some of the incorrect answers, the faithfulness of autoformalization remains a challenge for future work. 

\section*{Ethics Statement}

This work studies answer selection for mathematical reasoning using
public benchmarks (MATH-500, OlympiadBench, AMC-AIMO, AIME~2024) and
models; it involves no human subjects, no private
or personally identifiable data, and no sensitive content.
\textsc{LeanScribe} is fine-tuned on the openly available NuminaMath
dataset. The primary intended use is reducing the inference cost of
formal answer verification while improving the reliability, which lowers the energy and compute footprint of test-time scaling rather than increasing it. As we emphasize in Section~\ref{sec:prover}, a single Lean proof is not a
standalone correctness guarantee; practitioners should not treat the
formal signal as infallible certification, particularly in
high-stakes settings, since residual errors can arise from
mis-formalization. We see no foreseeable risk of harm specific to
this work beyond those general to mathematical reasoning systems.

\bibliography{custom}

\clearpage
\appendix

\section{Dataset Details}
\label{app:datasets}

\textbf{MATH-500} is the $500$-problem evaluation
subset~\citep{lightman2024prm800k} of MATH~\citep{hendrycks2021math}.
\textbf{OlympiadBench}~\citep{he2024olympiadbench}: we use the
English mathematics open-ended subset, restricted to single-answer
problems with \texttt{Numerical} or \texttt{Expression} answer
types, yielding $566$ problems; proof-only problems are excluded as
they fall outside the answer-construction setting
(Section~\ref{sec:intro}). \textbf{AMC-AIMO} is the AIMO validation
AMC set~\citep{aimo2024validation}: $83$ AMC12 problems from
2022--2023, extracted from the AoPS wiki and normalized to
integer-answer form. \textbf{AIME 2024} is the $30$-problem AIME
subset of the AIMO validation set~\citep{aimo2024validation}.

\section{Substitution Gate}
\label{app:gate}

The substitution gate referenced in Section~\ref{sec:formalization} filters
candidate positions through two structural checks:
\begin{enumerate}[leftmargin=*,itemsep=2pt]
    \item \textbf{Comment-aware filter.} Exclude positions inside Lean
    block comments (\texttt{/-~\ldots~-/}) and line comments
    (\texttt{--~\ldots}).
    \item \textbf{Theorem-body filter.} Restrict positions to the local block
    between the \texttt{theorem}/\texttt{lemma}/\texttt{example}
    keyword and the \texttt{:= by} proof marker, excluding imports,
    helper definitions, and content after the proof.
\end{enumerate}
A position survives the gate iff it passes both filters. The gate is
deterministic and stateless. We additionally short-circuit the base
candidate $a_b$ in the cascade: since the base statement $F_b$ type-checks by construction, we mark $a_b$ as verified without re-running the
substitution, avoiding spurious failures from non-round-tripping
edits. A per-candidate fallback further routes individual unsafe
candidates (e.g., natural-language labels) to \textsc{LeanScribe} even when
\textsc{R1} would otherwise apply; we ablate this in
Section~\ref{sec:ablations}.

\section{\textsc{LeanScribe} SFT Details}
\label{app:sft-details}

Each \textsc{LeanScribe} training example consists of:
\begin{itemize}[leftmargin=*,itemsep=2pt]
    \item \textbf{Input.} An informal problem $x$, a Lean statement
    $F_b$ that the autoformalizer produced for some verified base
    answer $a_b$, and the natural-language original answer string.
    \item \textbf{Output.} A JSON object specifying (i) the predicted
    \emph{local block} inside $F_b$ that encodes the answer, and
    (ii) a Python function \texttt{fill\_answer(answer: str) -> str}
    that, given any informal answer string, returns the Lean
    expression replacing that block.
\end{itemize}
The local block is required to occur exactly once in $F_b$ outside
comments and inside the theorem body; otherwise the example is
discarded during inference.

\paragraph{Lean-filtered expert iteration.} The corpus is built in
stages, each Lean-filtered. \emph{Phase~1} ($225$ examples) consists
of verified seed rewrites mined directly from \textsc{NuminaMath}
formalizations. \emph{Phase~2} ($699$ examples after strict
filtering) is generated by an intermediate \textsc{LeanScribe}
checkpoint under rejection sampling: the checkpoint produces several
block/function predictions per problem, and we keep only those whose
reconstructed statements pass an independent Lean type-check. We
additionally mine \emph{hard cases}---problems where direct
localization fails---with earlier checkpoints, again retaining only
Lean-passing rewrites ($3{,}262$ examples after de-duplication
against Phases~1--2). Merging these gives $4{,}186$ examples
($3{,}768$ train / $418$ test). The final \textsc{LeanScribe} is a
Qwen3-8B model LoRA-fine-tuned on this aggregated corpus, with rank~32,
$\alpha{=}64$, and dropout~$0.05$ on all attention and MLP
projections, for 3 epochs at learning rate $1\!\times\!10^{-4}$ in
bfloat16.
The \textsc{R1} disambiguation step (choosing among multiple
occurrences of the answer string) uses the same base model
\emph{without} the LoRA adapter, prompted to return the index of the
answer slot; only \textsc{LeanScribe} (\textsc{R2}) uses the
fine-tuned adapter.

\section{Drop-in Pipeline Setup}
\label{app:dropin}

The drop-in experiment (Section~\ref{sec:dropin}) embeds
base-and-edit into a \textsc{Fans}-style pipeline~\citep{fans2024}.
This setting differs from our main rank-based pipeline in three ways:
a single solver, Qwen2.5-Math-7B-Instruct, produces the $K{=}8$
candidate answers; an NL--FL consistency judge (QwQ-32B) filters
formalizations; and the final selector aggregates surviving candidates
by majority vote rather than by generator rank.

The full pipeline is proof-level. Candidate answers are converted into
formal statements using either independent Kimina formalization or
\textsc{Base}. \textsc{DeepSeek-Prover-V2} then generates proofs for
the resulting statements; QwQ-32B checks natural-language--formal
consistency; the Lean server verifies the generated proofs; and majority
vote is applied over candidates that pass both the proof and consistency
filters. Thus, in this section, ``verified candidates'' refers to
candidates whose generated proofs are accepted by Lean and whose formal
statements pass the NL--FL consistency judge, not merely statements that
type-check before proof search.

We hold the answer generator, prover, NL--FL judge, Lean proof verifier,
majority-vote selector, and candidate inputs fixed, and swap only the
formalization stage. Independent formalization is the \textsc{Fans}
default, while \textsc{Base} preserves the $K$ candidate slots and fills
them with base-and-edit formal statements. Therefore, the differences in
autoformalizer calls and final accuracy in Table~\ref{tab:dropin} are
attributable to the formalization strategy.

\section{Proof-Complete Criterion}
\label{app:proof-complete}

In the prover-swap ablation (Section~\ref{sec:prover}) we report
the \emph{proof-complete} rate---the fraction of statements for
which the prover returns a proof that Lean accepts with no remaining
\texttt{sorry} placeholder---rather than the raw pass rate.
\textsc{DeepSeek-Prover-V2} sometimes returns proofs that compile
but leave a \texttt{sorry}, which would admit a goal without proving
it; on MATH-500 its raw pass rate is $15.1\%$ but its proof-complete
rate is $7.3\%$. \textsc{Kimina-Prover} does not emit such proofs,
so the two rates coincide for it. We use proof-complete throughout
for a consistent comparison across provers.

\section{Fall-back Rate}
\label{app:abstentions}

On a small fraction of problems no candidate's autoformalization
type-checks in Lean, so neither independent formalization nor
\textsc{Base} has a verified statement to select from; for these we
fall back to the top-ranked candidate $a_1$
(Section~\ref{sec:formalization}). Table~\ref{tab:abstentions}
reports the per-dataset fall-back rate on the Gemini~3 Pro line.

\begin{table}[h]
\centering
\small
\setlength{\tabcolsep}{6pt}
\begin{tabular}{l r r r}
\toprule
Dataset & $n$ & Fall-backs & Rate \\
\midrule
MATH-500       & 500  & 6  & 1.2\% \\
OlympiadBench  & 566  & 12 & 2.1\% \\
AIME 2024      & 30   & 0  & 0.0\% \\
AMC-AIMO       & 83   & 0  & 0.0\% \\
\midrule
\textbf{Total} & \textbf{1179} & \textbf{18} & \textbf{1.5\%} \\
\bottomrule
\end{tabular}
\caption{Per-dataset fall-back rates (Gemini~$K{=}8$): the fraction
of problems where no candidate formalization type-checks. The rate is
identical for the baseline and \textsc{Base}, so it does not affect
their comparison.}
\label{tab:abstentions}
\end{table}

\section{Variance Reduction Detail}
\label{app:funnel}

The base-ablation of Section~\ref{sec:analysis-mech} (Table~\ref{tab:base-ablation})
excludes MATH-500: substituting alternative answers into its bases
produces statements whose Lean elaboration is prohibitively slow,
hitting the type-check timeout on nearly every candidate, so the
matched-set comparison there is run on OlympiadBench, AMC-AIMO, and
AIME~2024. This exclusion affects only the three-way base-ablation;
the main results (Table~\ref{tab:main}) and the factorization
(Table~\ref{tab:factorization}) do include MATH-500 and show the same
qualitative pattern there---\textsc{Base} raises
$\mathrm{Acc}\,|\,\mathrm{GT}@K$ while leaving $\mathrm{GT}@K$ nearly
unchanged---so we do not expect the shared-base effect to be specific
to the smaller datasets, though we cannot run the matched random-base
control on MATH-500 itself.

Formalizing once rather than $K$ times reduces the variance among a
problem's $K$ candidate statements. Measuring pairwise character
$n$-gram distance between the $K$ statements of each problem, \textsc{Base}
is $6.3\times$ more consistent than independent formalization,
stable across both solvers and all datasets. At the per-problem
level, the magnitude of variance reduction predicts the type-check
gain (Pearson $r{=}0.22$, $p{<}10^{-3}$, $n{=}2015$), and the gain
concentrates on problems the baseline struggles to formalize.
Importantly, the variance signal predicts the type-check gain but
\emph{not} the downstream prove-completion gain ($|r|{<}0.05$):
variance reduction acts at the formalization stage, raising how many
candidates---and how many \emph{correct} candidates---type-check and
reach the prover, rather than changing what the prover can close.

\section{Qualitative Formulation Examples}
\label{app:examples}

The examples below are drawn from MATH-500 (Gemini~3 Pro, $K{=}8$).
Each shows a single problem with its verified base statement $F_b$,
and then a second candidate answer $a_i$ formalized two ways: by
\textsc{Base} (editing only the answer block of $F_b$) and by
independent formalization (a fresh autoformalizer call). In every
case the \textsc{Base} edit type-checks ($\checkmark$) while the
independent statement does not ($\times$). The independent call,
re-deriving the whole statement from scratch, drifts in ways that
break the Lean check---dropping the theorem header, re-encoding the
goal, or changing the domain---whereas \textsc{Base} preserves the
well-typed structure of the shared base and rewrites only the answer
term. (\texttt{import Mathlib} headers are elided for space.)

\paragraph{Example 1: tuple answer split across a conjunction.}
\emph{Problem.} Find $(p,q,r)$ such that $x^3-3x^2+4x-1$ divides
$x^9+px^6+qx^3+r$. \emph{Base answer} $a_b=(6,31,-1)$; \emph{candidate}
$a_i=(-6,-31,1)$.
\begin{quote}\ttfamily\small
\textmd{\textit{$F_b$ ($\checkmark$):}}\\
open Polynomial\\
theorem thm \{p q r : \(\mathbb{R}\)\}\\
\phantom{xx}(h : (X\^{}3 - 3*X\^{}2 + 4*X - C 1 : \(\mathbb{R}\)[X])\\
\phantom{xxxx}$\mid$ (X\^{}9 + C p*X\^{}6 + C q*X\^{}3 + C r)) :\\
\phantom{xx}p = 6 $\wedge$ q = 31 $\wedge$ r = -1 := by sorry\\[4pt]
\textmd{\textit{\textsc{Base} $F_{b\to i}$ ($\checkmark$):}} \textmd{same body,}\\
\phantom{xx}p = -6 $\wedge$ q = -31 $\wedge$ r = 1 := by sorry\\[4pt]
\textmd{\textit{Independent $F_i$ ($\times$):}}\\
theorem (p q r : \(\mathbb{R}\))\\
\phantom{xx}(h : $\forall$ (z : \(\mathbb{C}\)), z\^{}3 - 3*z\^{}2 + 4*z - 1 = 0\\
\phantom{xxxx}$\to$ z\^{}9 + p*z\^{}6 + q*z\^{}3 + r = 0) :\\
\phantom{xx}p = -6 $\wedge$ q = -31 $\wedge$ r = 1 := by sorry
\end{quote}
\textsc{Base} edits the three scalar slots of the verified
divisibility statement. The independent call instead re-encodes
divisibility as a quantifier over complex roots and omits the
theorem name, and does not type-check.

\paragraph{Example 2: set-valued answer.}
\emph{Problem.} Find all real $x$ with
$(x-3)^3+(x-7)^3=(2x-10)^3$. \emph{Base answer} $a_b=\{3,5,7\}$;
\emph{candidate} $a_i=\{-3,7\}$.
\begin{quote}\ttfamily\small
\textmd{\textit{$F_b$ ($\checkmark$):}}\\
theorem thm :\\
\phantom{xx}\{x $\mid$ (x-3)\^{}3 + (x-7)\^{}3 = (2*x-10)\^{}3\}\\
\phantom{xxxx}= \{3, 5, 7\} := by sorry\\[4pt]
\textmd{\textit{\textsc{Base} $F_{b\to i}$ ($\checkmark$):}} \textmd{same body,} = \{-3, 7\}\\[4pt]
\textmd{\textit{Independent $F_i$ ($\times$):}}\\
theorem (x : \(\mathbb{R}\)) :\\
\phantom{xx}(x-3)\^{}3 + (x-7)\^{}3 = (2*x-10)\^{}3\\
\phantom{xxxx}$\leftrightarrow$ x = -3 $\vee$ x = 7 := by sorry
\end{quote}
\textsc{Base} keeps the set-equality form and edits the solution set.
The independent call rewrites the same answer as a biconditional over
a free variable (and drops the header), changing the proposition's
shape; it does not type-check.

\paragraph{Example 3: domain divergence ($\mathbb{Z}$ vs.\ $\mathbb{R}$).}
\emph{Problem.} For $f:\mathbb{R}\to\mathbb{R}$ with
$f(x)+f(y)=f(x{+}y)-xy-1$ and $f(1)=1$, find the integer fixed
points. \emph{Base answer} $a_b=\{-2,1\}$; \emph{candidate}
$a_i=\{-1,1\}$.
\begin{quote}\ttfamily\small
\textmd{\textit{$F_b$ ($\checkmark$):}}\\
theorem thm \{f : \(\mathbb{R}\) $\to$ \(\mathbb{R}\)\}\\
\phantom{xx}(h : $\forall$ x y, f x + f y = f (x+y) - x*y - 1)\\
\phantom{xx}(h' : f 1 = 1) :\\
\phantom{xx}\{n $\mid$ $\exists$ m : \(\mathbb{Z}\), n = m $\wedge$ f n = n\} = \{-2, 1\}\\
\phantom{xx}:= by sorry\\[4pt]
\textmd{\textit{\textsc{Base} $F_{b\to i}$ ($\checkmark$):}} \textmd{same body,} = \{-1, 1\}\\[4pt]
\textmd{\textit{Independent $F_i$ ($\times$):}}\\
theorem (f : \(\mathbb{R}\) $\to$ \(\mathbb{R}\)) (h : \ldots) (h' : f 1 = 1) :\\
\phantom{xx}\{n : \(\mathbb{Z}\) $\mid$ f n = n\} = \{-1, 1\} := by sorry
\end{quote}
\textsc{Base}'s base encodes the fixed-point set over
$\mathbb{R}$ with an integrality constraint. The independent call
re-types the set over $\mathbb{Z}$ and applies
$f:\mathbb{R}\to\mathbb{R}$ to an integer index, which is ill-typed;
it does not type-check.

\paragraph{Summary.}
Across these cases the same answer is well-typed under \textsc{Base}
but not under independent formalization---not because the answer is
intrinsically hard to formalize, but because re-formalizing from
scratch introduces structural and typing variance that the shared
verified base eliminates. This is the per-example face of the
variance reduction quantified in
Appendix~\ref{app:funnel}.

\end{document}